\documentclass{llncs}

\usepackage{graphicx}
\usepackage{hyperref}
\usepackage{color}
\usepackage{hyphenat}
\usepackage{textcomp}
\usepackage[T1]{fontenc}
\usepackage[utf8]{inputenc}
\usepackage{epigraph}
\renewcommand{\refname}{References}

\begin{document}
%*****************************VITELLONE*********************************************************
\title{A giant with feet of clay: on the validity of the data that feed machine learning in medicine\thanks{In case you like to cite this article in your work, would you please consider the opportunity to mention the version that we published on a book by Springer? This can be referred as follows:  Cabitza, F., Ciucci, D., Rasoini, R.: A giant with feet of clay: on the validity of the data that feed machine learning in medicine. In: Cabitza, F., Magni, M., Batini, C. (eds.) Organizing for the Digital World, Lecture Notes in Information Systems and Organisation, vol. 28, chap. 2, pp. 113–128. Springer, The address of the publisher (2018) ISBN: 978-3-319-90503-7}}
\titlerunning{giant with feet of clay} % abbreviated title (for running head)
%                   also used for the TOC unless
%                   \toctitle is used
%
\author{Federico Cabitza\inst{1,2} \and Davide Ciucci\inst{1} \and Raffaele Rasoini\inst{3}}
\authorrunning{Cabitza et al.} % abbreviated author list (for running head)
\tocauthor{cabitza, Ciucci, Rasoini}
\institute{Universit\'{a} degli Studi di Milano-Bicocca, Milan, Italy\\
\email{cabitza,ciucci@disco.unimib.it}
\and
IRCCS Istituto Ortopedico Galeazzi, Milan, Italy\\
\and 
IFCA Istituto Fiorentino di Cura e Assistenza, Florence, Italy
\email{raffaele.rasoini@tiscali.it}
}

\maketitle       % typeset the title of the contribution

\begin{abstract}
This paper considers the use of Machine Learning (ML) in medicine by focusing on the main problem that this computational approach has been aimed at solving or at least minimizing: uncertainty. To this aim, we point out how uncertainty is so ingrained in medicine that it biases also the representation of clinical phenomena, that is the very input of ML models, thus undermining the clinical significance of their output. Recognizing this can motivate both medical doctors, in taking more responsibility in the development and use of these decision aids, and the researchers, in pursuing different ways to assess the value of these systems. In so doing, both designers and users could take this intrinsic characteristic of medicine more seriously and consider 
alternative approaches that do not ``sweep uncertainty under the rug'' within an objectivist fiction, which everyone can come up by believing as true.  

\keywords{Decision Support Systems, Machine Learning, Information Bias, Algorithmic Bias}
\end{abstract}
\section{Motivations and Background}
\label{sec:intro}

It is a truism to say that uncertainty permeates contemporary medicine~--~not much differently than it has always done~--~as it has also been confirmed by extensive studies in the field of sociology and medicine itself (e.g.,~\cite{fox2000medical,rosenfeld2003uncertainty,simpkin2016tolerating,hatch2017uncertainty}). 
However, uncertainty is a broad term, which encompasses many types of shortcomings of human knowledge. 
We cope with some form of uncertainty when we cannot pinpoint a phenomenon exactly or when we cannot measure it precisely (i.e., approximation, inaccuracy); when we do not possess a complete account of a case (incompleteness, inadequacy); when we cannot predict what it 
will come next (unpredictability for randomness or excessive complexity); when our observations seem to contradict each other (inconsistency, ambiguity); and, more generally, when we are not confident of what we know. In clinical practice, all of these phenomena occur on a daily basis, several times. Greenhalgh~\cite{greenhalgh2013uncertainty} has recently proposed a taxonomy of uncertainty in medicine: she distinguishes between uncertainty about the best available evidence (will it be sound? will it be generalizable?); about the story of patients (are they sincere? are they reliable?); about case-specific decisions (e.g., what best to do in the circumstances?); and about social or soft skills (e.g., how best to communicate and collaborate with colleagues?). 

More in general, medical doctors can be uncertain on almost every aspect of their practice: on how to classify patients' conditions (diagnostic uncertainty); why and how patients develop diseases (pathophysiological u.); what treatment will be more appropriate for them (therapeutic u.); whether they will recover with or without a specific treatment (prognostic u.), and so on. 
In this picture, technology has often been proposed~--~and seen~--~as a solution. In the words by Reiser~\cite[p. 18]{reiser1984machine}: 
\textit{From the beginning of their introduction in the mid-nineteenth century, automated machines that generated results in objective formats [...] were thought capable of purging from health care the distortions of subjective human opinion [and] to produce facts free of personal bias, and thus to reduce the uncertainty associated with human choice.}

Also computing technology has been proposed to address all of the above areas of uncertainty~--~to either try to control or minimize it: the first computational support, what was then called a rule-based expert system, was introduced more than 40 years ago to propose a ``quantification scheme which attempts to model the inexact reasoning processes of medical experts''~\cite{shortliffe1975model}.

After the introduction of many and different computational systems~\cite{schwartz_artificial_1987}, a new class of applications has recently emerged in the health care debate: the \emph{decision support systems} (DSS) that embed predictive models that have been developed by means of \emph{machine learning} (ML) methods and techniques. These systems, which for the sake of brevity we will call ML-DSS, have recently raised a strong interest among the medical practitioners of almost every corner of the world in virtue of their high accuracy at an unprecedented extent~\cite{esteva2017dermatologist,gulshan2016development}, in some cases even allegedly capable of outperforming human experts (e.g.,~\cite{kooi2017,wu2016classifier}). This is reflected by the stance of influential commentators and medical experts that have recently shared their thoughts from some of the most impacted journals of the medical community (e.g.~\cite{deo2015machine,obermeyer2016predicting,darcy2016machine,jha2016adapting,leachman2017medicine,gillies2017viewpoint}, just to cite a few). These voices do not clearly indulge in techno-optimistic claims and do not refrain from offering some words of caution; however, the recent successes of ML-DSS in medical imaging (e.g., radiography, computed tomography, ultrasonography, magnetic resonance imaging, retinal fundus photography, skin images) pose the issue of how these systems and their improved versions, which will likely outperform human diagnosticians, will impact some medical professions like radiologists and pathologists~\cite{obermeyer2016predicting,chockley2016end,jha2016adapting}, and health care in general~\cite{deo2015machine,darcy2016machine}.
In regard to this impact, two elements should be object of further scholarly interest and research, which are bound together by a feedback loop making their mutual influence subtle but yet hard to pinpoint. First: how ML-DSS can bias human interpretation and decision, or \emph{automation bias}. Second: how human interpretation and classification can bias the ML-DSS performance, or \emph{information bias}. While the former case is still neglected but some first studies are shedding light on it~\cite{parasuraman2010complacency,goddard2012automation,cabitza2017breeding}, in this paper we will focus on the latter case, which is almost completely ignored, especially by the computer scientists and designers of ML models. 
Nevertheless information bias, which we will define in the next section, regards the quality\footnote{This is a vague term: here we mean data quality mainly in terms of accuracy and validity.} of both the training and input data of ML-DSS, and hence has got the potential to undermine the reliability of the output of ML-DSS. Our point is that a renovated awareness of the irreducible nature of the uncertainty of medical phenomena, even in regard to their plain representation into medical data, can help put in the right perspective the current potential of ML-DSS, and motivate the exploration of alternative ways to conceive them and validate their indications, as we will outline in the last sections of this contribution.

\section{Seeing what we are trained to see}
\label{sec:infobias}

Before considering information bias from a medical perspective, let us recall what a ML predictive model is. A predictive model, no exception those developed with a machine learning approach\footnote{In what follows we introduce the concept of ML predictive model with reference to supervised discriminative (or classification) models. Although machine learning models can be of various kinds, the above models are by far the most frequently used in medicine. What we say for this particular kind of ML is easily applicable to other forms of ML with just minor changes in wording.}, are \emph{functionally relational} models that bind input data to one category out of a set of predefined ones (most of the times encompassing only two options, like positive/negative). This latter category is the output (or prediction) of the model. 

To this aim, the model is progressively fine-tuned on the basis of what ML experts call \emph{experience}~\cite[p. 2]{mitchell1997machine}, that is input data that have been already classified in terms of a specific category. In the case of medical classification (for both diagnostic or prognostic aims) the above ``experience'' is a set of cases that have been already represented properly and classified ``correctly'' \emph{according to some gold standard method}. In so doing, the machine can learn the model, that is the hidden structural relationship between the features of the cases (as long as these are all represented in terms of the same attributes and characteristics), and hence the correct interpretation of each case. Grounding on this model, the ML-DSS can ``predict'' the right category or label when fed with new cases, as long as these are sufficiently similar to those ones with which it has been trained. 

As widely known, ML encompasses many families of models and, within each family, every model can differ from the others for the countless combinations of its various internal parameters. The reason why different models perform differently on the same data is that they represent them differently and make different assumptions on their intrinsic regularities: in short, they are all differently \emph{biased}. This term is often used in the ML community, not always with precision or consistency, to account for the many factors that condition a model in predicting the output. The preference for simpler models, or the Occam's razor, is an example of active bias in model optimization and selection.

\begin{figure*}[tb]
\centering
 \includegraphics[width=.8\textwidth]{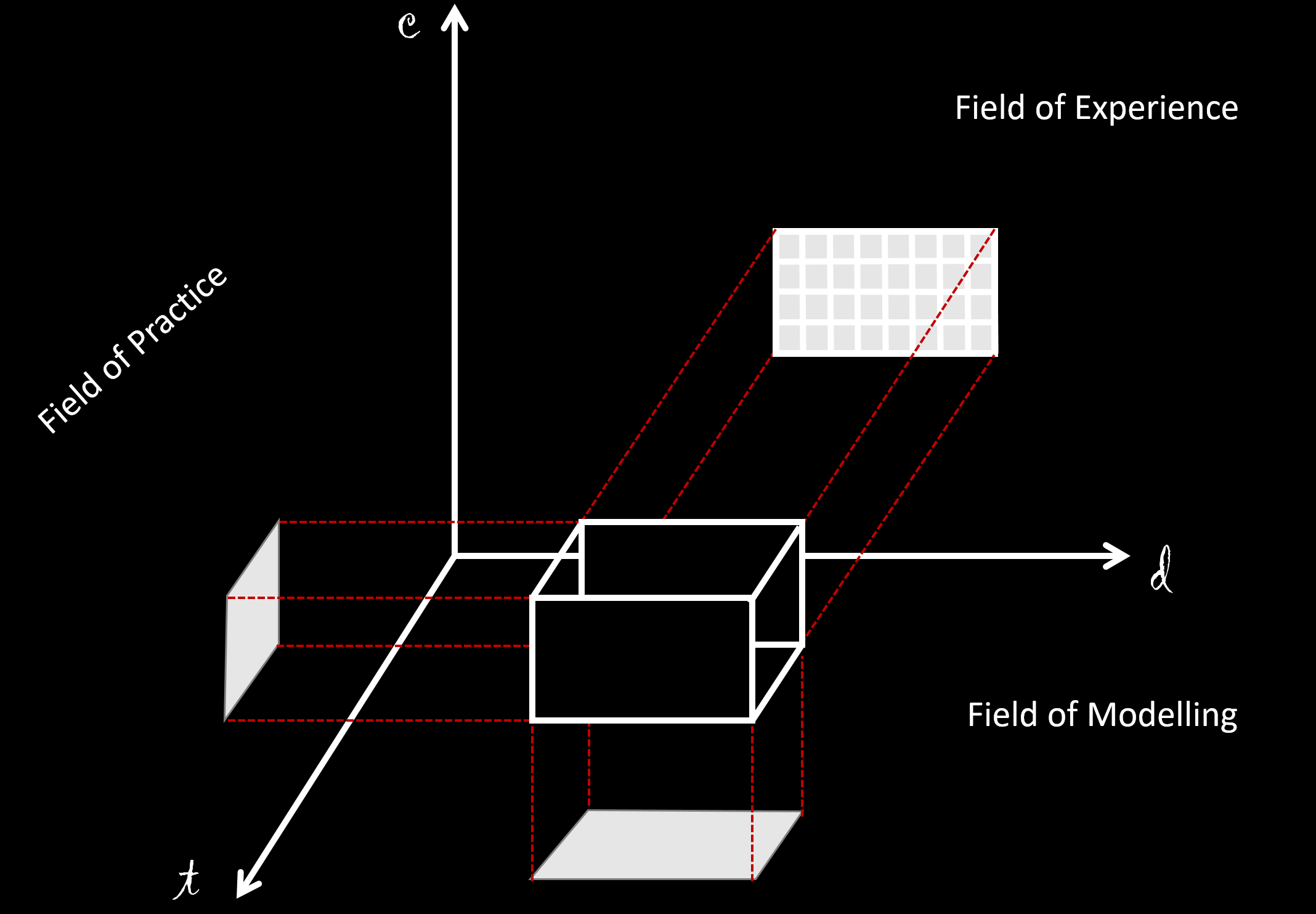}
\caption{An illustration of the fact that we represent only a small portion of reality. The three axes represent the cases (C), the dimensions (D) and time (T). The experience is represented in tabular form in the Experience field (the C-D plane). Dimensions (or columns in the tabular form) can be added, deleted or renamed over time (plane of modeling, D-T). Cases (or rows in tabular form) can be added, deleted or, more frequently, modified (i.e., the output and effect of any data-oriented practice).}
\label{fig:representation}   
\end{figure*}

Once again Mitchell has the merit to have given a simple yet clear definition of bias in the ML discourse: this is \emph{any basis for making a specific prediction over another, other than strict consistency with the observed training instances}~\cite{mitchell1980need}. The most general kind of bias is \emph{inductive} bias (also called learning bias\footnote{The name comes from the main assumption of any learning process, which was first acknowledged by Hume in the 1700s: ``we suppose, but are never able to prove, that there must be a resemblance betwixt those objects, of which we have had experience, and those which lie beyond the reach of our discovery'' (from A Treatise of Human Nature, Book I, Part III,
Section VI, 1740).}), whose main components are \emph{representational bias} (also called language bias) and \emph{algorithmic bias} (also called procedural bias)~\cite{gordon1995evaluation}. The latter bias regards the number of assumptions and heuristics by which a model is refined, that is by which the optimization procedure traverses the space of all possible hypothetical models and finds the fittest to the available experience.
Representational bias is somehow wider and covers different aspects that are nevertheless connected regarding what structure the model family assumes to adequately capture regularities in the data (e.g., a decision tree), what constitutes a complete representation of the above space of all possible fitting models and, last but not the least, any assumption in representing entities, states and events in the reality of interest, that is any modeling choice that is translated in considering certain features and attributes and not other, in their types and their range of values. Figure~\ref{fig:representation} illustrates (in spatial terms) the extent of any representation of the reality of interest. 

Thus we come to the main issue at stake in this paper: how much valid and reliable is the \emph{representation of the above ``experience''} on which ML-DSS learn their predictive model? Two main biases weigh on this question: the above representational bias and the so called information bias~\cite{althubaiti2016information}. By information bias we intend a collective name for all the human biases that distort the data on which a decision maker (or a computational decision support) relies on, and that account for the validity of data, that is the extent these represent what they are supposed to represent accurately. 
These two biases, and the related phenomenon of \emph{information variability}, all concur in undermining the extent we can be certain of the available data, and hence of the predictions ML models can infer from them. 

The relationship between representational bias and information bias is mutually conditioning and so tight that distinguishing between them could be improper at a high level of description\footnote{If we could choose the terminology that we deem more accurate, we would propose to denote as representational bias the general phenomenon, which encompasses both a modeling bias (regarding the conceptual model of the data) and an information bias (regarding the description of the reality in terms of that conceptual model). However, this choice would likely produce more confusion in the specialist literature, which is already cluttered with different forms of biases and their effects.}, as both regard the shortcomings that are inherent to any classifying taxonomy and measurement scales\footnote{These shortcomings would deserve a study of its own. In a famous work, Star and Bowker~\cite[p. 69]{bowker2000sorting} hinted at some of these inadequacies, which include: temporal rigidity; a one-size-fits-all nature in regard to meaning and implications; and, as also discussed in~\cite{timmermans2010gold}, the reflection of disciplinary interests, agendas and priorities.}. To borrow terms from data-base conceptual modeling, we could see representational bias as affecting the creation of the \emph{schema} of our experience representation: the choice of what attributes and value ranges designers deem relevant to adopt, the standard classifications that the designers considered adequate to pinpoint the aspects of the reality of interest: in short, any bias in drawing an empty table. On the other hand, information bias affects the situated activities of filling in this table, the ``instancing'' the above schema, and the ``projection'' of the observed phenomena into the available data structure. In some ways information bias can be seen as the  ``grounding'' of representational bias, or the \emph{user side} of it. 
%Figure~\ref{fig:problems} illustrates the main problems regarding representational and information bias with respect to the metaphor adopted in Figure~\ref{fig:representation}.

% \begin{figure*}[tb]
% \centering
%  \includegraphics[width=.9\textwidth]{problems.png}
% \caption{The main threats to an accurate and complete representation of medical cases for the aim of automated prediction.}
% \label{fig:problems}   
% \end{figure*}

Since ML specialists tend to overlook this side, in the next section we will characterize this concept further, and discuss its role with respect to uncertainty in medical decision making.

%FC da recuperare forse...
%What distorts the representation of the single raters' (that is the physicians') judgment and hence affects the act of recording it (that is the choice of any symbolic label, code, number or any written sign to represent the continuous phenomenon that the raters have observed and examined), is called \emph{information bias}~\cite{althubaiti2016information}. 

\section{Information bias, the open secret of medical records}
\label{sec:infobiasmed}

Information bias\footnote{While biases are, strictly speaking, mental prejudices, idiosyncratic perceptions and cognitive behaviors producing an either impairing and distorting effect, here we rather intend the effect (by metonymy), that is the ``error'' in the data recorded and the decisions taken caused by the bias itself.} accounts for any factor that could undermine the validity of a representation, that is the extent it truly represents one or more aspects of the reality of interest: in short what undermines the validity of the available data. This general bias can take various forms including, most manuals concur, measurement error, misclassification and miscoding. However, this bias should \emph{not} be only associated with errors and mistakes by the ``data producers'' (in our case medical doctors, nurses and patients)  due to either negligence, fraud, incompetence, incomprehension or inexperience. %Rather it encompasses phenomena that are often neglected, or poorly studied: 

\begin{figure*}[tb]
\centering
 \includegraphics[width=\textwidth]{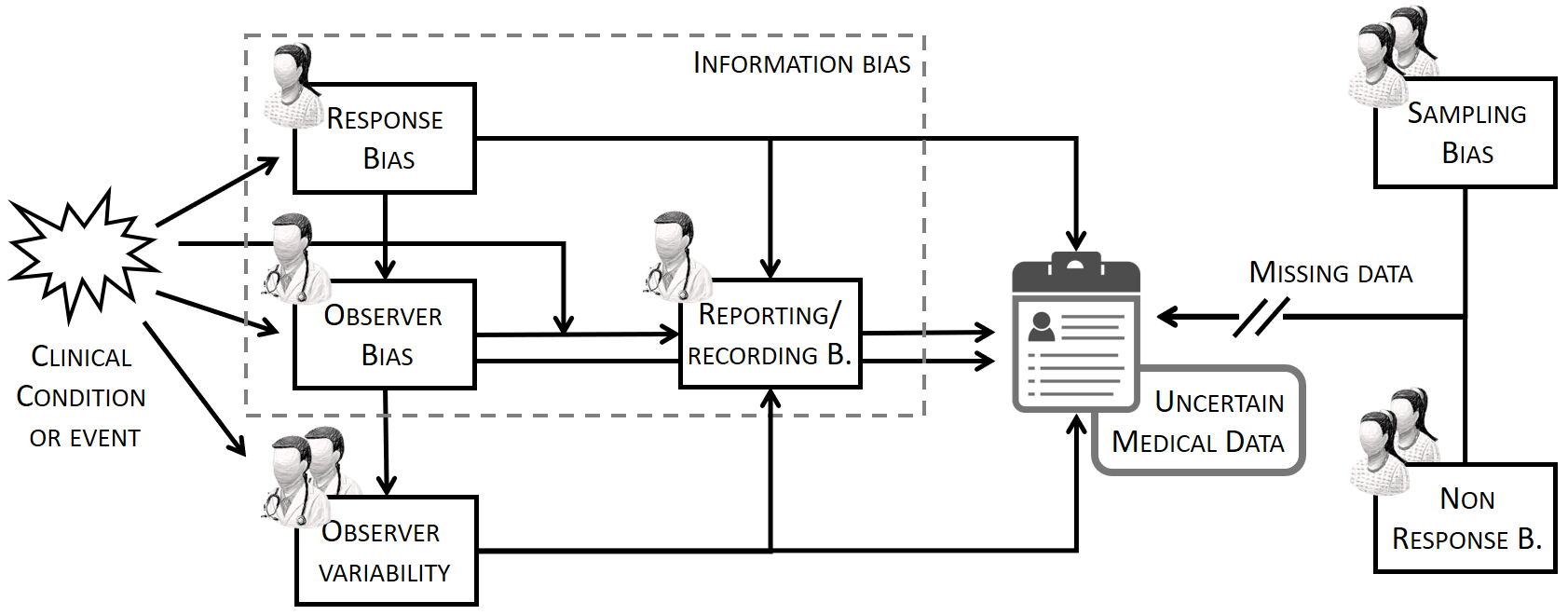}
\caption{The main biases affecting the validity and reliability of medical data. Sampling and Nonresponse biases are indicated to account for the lack of information that, if present, could reduce uncertainty of representation.}
\label{fig:schema}   
\end{figure*}

In medicine, information bias can be due to both patients and care givers in different but intertwined ways (see Figure~\ref{fig:schema}). Patients can (often unaware) contribute in terms of \emph{response bias}. This bias occurs when patients either exaggerate or understate their conditions (for many reasons and often in good faith), or whenever they intentionally suppress some information (e.g., like in case of sexually transmitted diseases or drug history for the related social stigma) or distort it (e.g., to avoid legal consequences); or when their recall is limited or flawed, or simply because they do not understand what physicians ask them or they aim to respond how (they believe) physicians expect them to (cf. the particular kind of response bias known as ``social desirability bias''). A large extent of response bias can be related to the inability of doctors to get confidence of the respondent~\cite{holt2017}. As an example, a recent study~\cite{shafiq2016patient} focused on the degree of concordance agreement between patients and cardiologists in regard to the presence of angina pectoris and its frequency. The study showed that when patients reported monthly angina symptoms, cardiologists diagnoses were compatible with patient reported symptoms only 17\% of time, while among patients who reported more frequent, i.e. daily/weekly angina symptoms, approximately one quarter of them were noted as having no angina by cardiologists. 

Besides the above mentioned condescending bias, it is well known that patients can exhibit behaviors (cf. Hawthorne effect), or even detectable modifications of some physiological parameter (e.g., blood pressure in what is also known as `white-coat effect') when they are under examination in a clinical setting that they would not exhibit in other settings.

Response bias is tightly associated with the main source of bias from the physicians' side: \emph{observer bias}. This occurs whenever the observer affects a measure, its accuracy and completeness, which clearly should not depend on the act of observing and measuring itself. Physicians introduce observer bias in their data due to both perceptual, cognitive and behavioral traits, weaknesses or just ``bad habits''; this bias also occurs whenever they, ``unwittingly (or even intentionally) exercise more care about one type of response or measurement than others, for instance, those supporting a particular hypothesis versus those opposing the hypothesis''~\cite{holt2017} (cf. confirmation bias).

Those who observe a clinical condition are often those who report it in the medical record. In this case observer bias can blur with what it is denoted as either recording, reporting, or coding bias.
In particular, this latter distorting factor can be traced back to many causes, from the least common and most poorly studied, like digit preference and conflictual coding, to the most pervasive ones, like the intrinsic inadequacies of any classification schema. 
Conflictual coding can affect the accuracy and completeness of medical data when proper reporting clashes with the personal interests of those who are supposed to document a clinical condition (like in the case of blood pressure recordings within a quality and outcome assessment framework of incentives~\cite{carey2009blood}). Digit preference occurs when measurements are more frequently recorded ending with 0 or 5, or as results of arbitrary rounding off. In~\cite{hayes2008terminal} the authors observed a much larger occurrence of these two digits in renal cell carcinoma measurement (p$<$0.0001) and concluded that this recording behavior could affect the determination of tumour stage, ``with resulting consequences in regard to prognosis and patient management''. Moreover, coding variability that leads to a lack of reliability can happen even when instructions on how to proper code are well known: a study~\cite{andrews2007variation} compared consistency of coding supposedly clean and high-quality data-sources like clinical research forms from observational studies among three professional coders, each using the same terminology and with the same instructions. All three coders agreed on the same core concept 33\% of the time; two of the three coders selected the same core concept 44\% of the time; and, no agreement among all three was found 23\% of the time. Moreover, no significant level of agreement beyond that due to chance was found among the experts. The same conclusion was drawn in a study where three shoulder specialists tried to evaluate 50 patients with shoulder instability and classified their condition using one of the 16 ICD-9 codes for shoulder instability~\cite{throckmorton2009intraobserver}.

%FC qui ci starebbe bene qualche esempio della inadeguatezza della ICD9-CM, come quella JAMA che ti ho segnalato 

Last, but not least, information bias in medicine can also be traced back to some sort of \emph{intrinsic ambiguity} of the medical conditions being documented, due to either their instability over time, or to variability across subjects and across observers. %An example of the former type of variability regards the reporting of blood pressure (BP): a study~\cite{huang1999stability} indicates ``significant differences between [\ldots] three subsequent BP measurements [taken with a 5 minute interval] (P$<$0.001), with much smaller differences between the second and third measures'' and a magnitude of fluctuation of the BP of 5 mmHg or less for most patients\footnote{Accordingly the BP screening protocol mentioned in the above paper recommended to report BP by averaging the second and third measure.}. 
A noteworthy example of this sort of ambiguity can be found in a recent study by Dharmarajan and colleagues~\cite{dharmarajan2016treatment}. This study focused on elderly people diagnosed (at hospital admission) with one of the following conditions: pneumonia, chronic obstructive pulmonary disease, or heart failure. These are three common conditions of the elderly that are responsible for breathlessness and other warning symptoms usually requiring hospital admission. The authors showed that patients regularly received, during hospital stay, concurrent treatment for two or more of the above cardiopulmonary conditions and not only for the main diagnosis identified at hospital admission. This exemplifies the fact that in real-world clinical practice, patients' clinical pictures are often blurry and not capable of being associated with clear-cut labels as expressed instead in textbooks and clinical practice guidelines. Indeed, even common clinical syndromes have disease presentations that often fall in-between traditional diagnostic categories. The common and relevant overlap of medical treatments as in the case mentioned above highlights the intrinsic ambiguity of clinical phenomena and the downstream uncertainty that medical doctors face in choosing what they deem a single right therapeutic strategy for a specific disorder.  

\begin{figure*}[tb]
\centering
 \includegraphics[width=\textwidth]{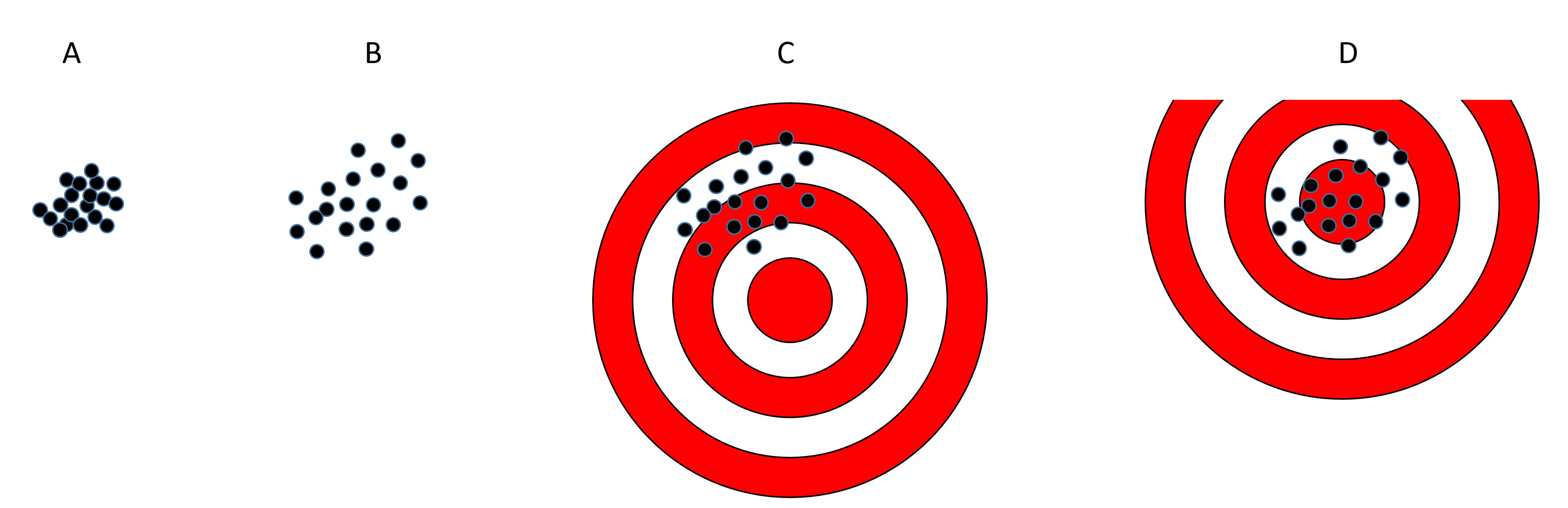}
\caption{Differences between validity and reliability. Each point is a multi-dimensional representation of a case (or observation). \textit{A} depicts the case of likely reliable data (though nothing can be said of their validity lacking a reference truth); \textit{B} depicts the case of less reliable data than \textit{A} (e.g., because variance is larger, raters could disagree among each other). \textit{C} depicts the same data of \textit{B} but the ground truth allows to evaluate bias (i.e., the offset with respect to the bullseye) and hence validity. \textit{D} depicts the same data as in \textit{B} and \textit{C}, but these data result now to be more valid because a different ground truth is taken as reference. Notably in medicine, differently from many engineering disciplines, also the ground truth can ``move'' according to the gold standard taken as reference (see Section~\ref{sec:ghost}).}
\label{fig:validity}   
\end{figure*}

Unfortunately, all of the kinds of variability and biases mentioned above cannot be conclusively addressed by improving the accuracy of any measurement tool, or by any other contrivance conceived from the engineering standpoint. 
%\emph{Observer variability} is an even more interesting and powerful phenomenon in affecting the quality of medical data. This kind of variability is mainly due to disagreement among multiple observers or among each observer with themselves on different occasions. 
Moreover, the extent these biases are expressed in a clinical setting varies a lot: although they look as abstract and general categories, biases are always exhibited by someone in particular, they are highly situated, and depend on personal skills, like clinical perspicuity, life-long acquired competencies, and contingent workloads. Since the impact of personal biases are difficult (if not impossible) to prevent, medical organizations try to minimize them with redundancy of effort, like relying on double checking and on second (or multiple) opinion. However, multiple opinions are both a resource to fight biases (by averaging multiple observations and measures), and, paradoxically enough, a source of low reliability and further uncertainty (`quot capita, tot sententiae'). 

\begin{figure*}[tbh]
\centering
 \includegraphics[width=.85\textwidth]{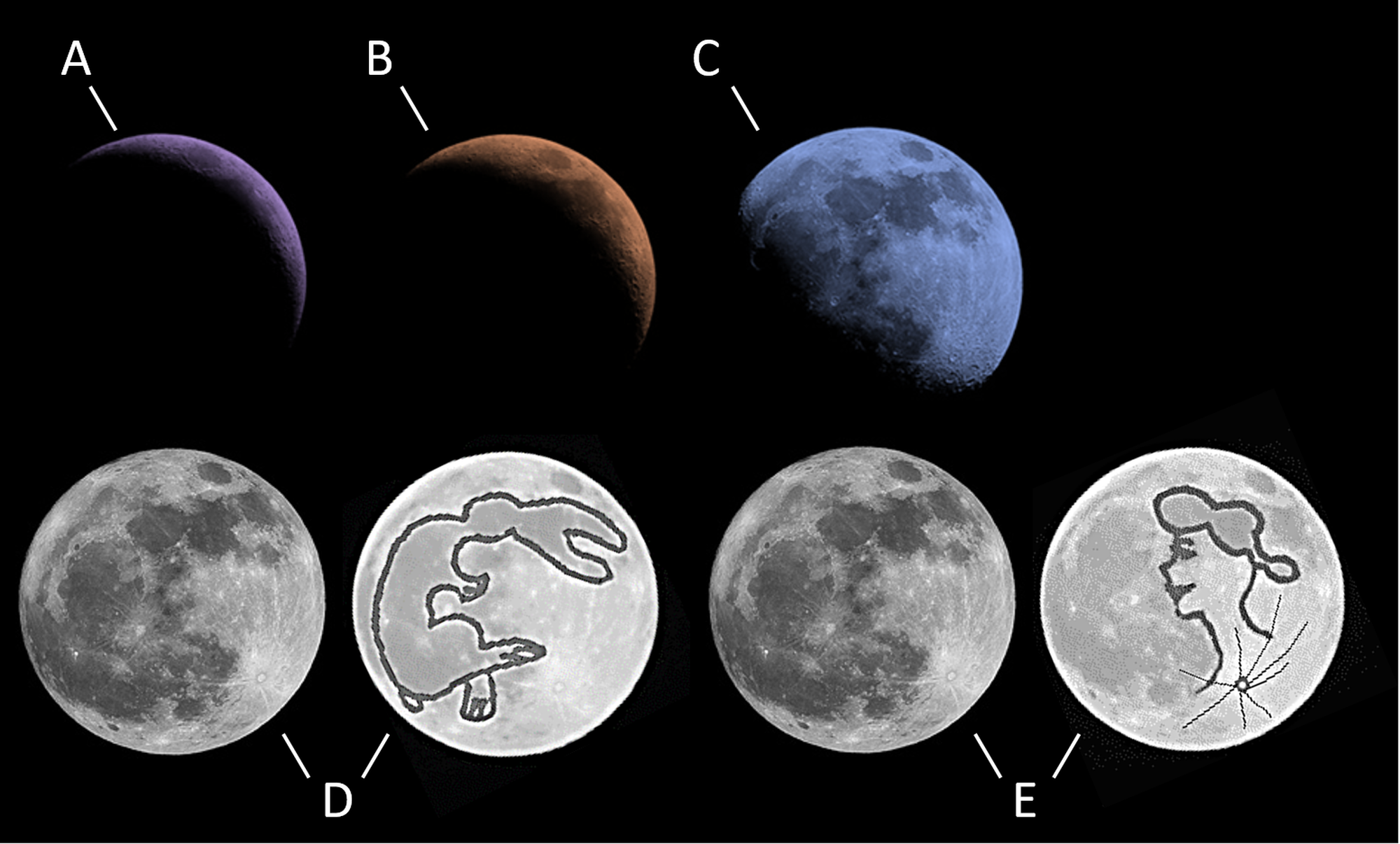}
\caption[A, B and C illustrate a simple case of homomodality, where different appearances of the moon are commonly classified with a single label: \texttt{waxing moon}. Moreover an English speaking observer would denote A as purple, B as brown and C as blue. An observer from the Ivory Coast speaking the local language Wob\'{e} would denote all of them as `Kpe'. D and E show a case of symmodality, where the same observable phenomenon (full moon) can be associated with two different labels: rabbit and lady]{A, B and C illustrate a simple case of homomodality, where different appearances of the moon are commonly classified with a single label: \texttt{waxing moon}. Moreover an English speaking observer would denote A as purple, B as brown and C as blue. An observer from the Ivory Coast speaking the local language Wob\'{e} would denote all of them as `Kpe'. D and E show a case of symmodality, where the same observable phenomenon (full moon) can be associated with two different labels: \texttt{rabbit} and \texttt{lady}\footnotemark.}
\label{fig:moon}   
\end{figure*}

Indeed, when more than one physician are supposed to determine the presence of a sign, make a diagnosis, or assess the severity of a condition, \emph{observer variability} (see Figure~\ref{fig:schema}) is introduced to account for the discrepancies in their opinions and for any difference in considering the same conditions. Observer variability has to do with the reliability of the judgment of so called medical ``raters'', and with the agreement that these latter ones reach independently of each other when they either measure, classify or interpret the \emph{same} phenomenon (e.g. an electrocardiogram, a radiography, a pathological sample, etc.) to make a decision, mainly a diagnosis (see Figure~\ref{fig:moon} for an example taking the moon as the common phenomenon). Observer variability affects the reliability of data, that is the extent the representations of the same or supposedly similar phenomena are equivalent or at least consistent with each other. Both validity and reliability jointly relate to the overall trustworthiness of data but the former is much more difficult to assess if there is a lack of a reference truth of which one is certain or sufficiently confident about (see Figure~\ref{fig:validity}).

\footnotetext{Three further remarks: Our color-related example regards only different ways to categorize colors, not to perceive them. Incidentally, the incidence of color blindness among African people is approximately \emph{half} the incidence among Caucasians. Moreover, the fact that recognizing different shapes in the same scene can be traced back to the known phenomenon of psychological pareidolia is irrelevant to our aims. For us this is just a common-sense example of possible multiple interpretations of the same phenomenon. Lastly, the careful reader will have noticed that the words \emph{moon}, \emph{month}, \emph{measure} and\ldots \emph{medicine} all derive from the same root *me-, Proto-IndoEuropean for ``to measure (appropriately)''.}

In medicine, not only multiple raters could classify the same phenomenon in different ways, but also the same doctor can disagree with herself, examining the same case after a certain amount of time, or in different environment conditions (e.g., with respect to workload, interruption rate, work shift). In the first case, researchers speak of \emph{inter-rater agreement}; in the latter case of \emph{intra-rater agreement}. Examples abound. For instance, in~\cite{jewett1992potential} the authors report the case of plain abdominal radiographies submitted to 3 different radiologists to detect the presence or absence of residual stone fragments: differences among radiologists were found in the 52\% of the reports, and even by the same radiologist rereading the films 24\% of the time. 
%RR aggiungo esempio 
As another example, we mention the case of heart gallop rhythm detection on cardiac auscultation, originally described almost 200 years ago~\cite{o1939gallop}, an objective sign that is associated with the serious clinical syndrome of heart failure. Despite the long lasting experience among clinicians all over the world with this warning objective sign (also due to the widespread availability and use of stethoscopes for cardiac auscultation among all medical disciplines), it has been shown~\cite{lok1998accuracy} that the agreement between expert or unexperienced observers and the phonocardiographic gold standard in the correct identification of gallop rhythm is very poor, with overall inter-rater agreement resulting little better than chance alone.
%FC non si può anticipare il discorso del kappa, non abbiamo ancora introdotto l'argomento, basta quello che si dice sopra.
%(Kappa of 0.05 and of 0.18 for the detection of the two different components of the gallop rhythm).  

%\textcolor{red} {As one of multiple possible examples, we cite the case of diagnosis of retinopathy in premature infants, a disorder related to an abnormal vascular development of the retina, observed in prematurity, and a major cause of childhood blindness. A multicenter, prospective study on 260 digital images of premature infants was performed to investigate intra- and inter-observer reliability among 7 \emph{experts} about diagnosis of retinopathy of prematurity~\cite{gschliesser2015inter}. Results showed both an inadequate inter-observer agreement among experts with resulting Fleiss' k values between 0.24 and 0.41 (according to diagnosis of different features in retinal images), and, as expected, a slightly higher (albeit still inadequate) intra-observer agreement for the same features with resulting Cohen's k between 0.47 and 0.63. These data highlight the existing high level of uncertainty and subjectivity surrounding the diagnosis of retinopathy of prematurity even among skilled experts (pediatric ophthalmology consultants specializing in retinopathy of prematurity and working in different University Hospitals in Europe), and thus the risk of observer bias about a medical condition whose prompt and early recognition may prevent blindness through laser photocoagulation treatment}.  
To account for the extent the majority decision (in case of multiple raters) or the category chosen more frequently can be considered reliable, and hence ``true'', the so called \emph{inter-rater reliability} (IRR) is measured, as we will see in the next section. 

\section{Between Gold Standards and Ghost Standards} 
\label{sec:ghost}

A ``gold standard'' (or with a less evocative but more correct expression ``criterion standard'') is a reference method to ascertain medical truth or evidence. By `reference' here we mean the `best one' under reasonable conditions, that is the method that `by definition' is capable to pose the so called ``ground truth''\footnote{This is an expression borrowed from cartography, where it indicates information that is acquired by direct observation (as opposed to inference, intelligence, reports, maps, etc.) in an actual field check at a location.} for any practical aim, including a scientific research and ML-DSS development. However, the degree of truth that a gold standard usually reaches is far from resembling an accurate, unambiguous and unique representation of medical facts that computer scientists long for their ``ground-truthing'', i.e., the process of gathering objective data to train a ML model. 

In fact there are many possible gold standards, even for the same disorder, ranging from autopsy (post-mortem) examination (in many cases
the strongest source) to the opinion of independent raters that could not receive strict indications on how to code what they observe and interpret (that is a much more common situation). However even in regard to those tests that are usually considered the most reliable and definitive gold standards, like post-mortem, histological and genetic examinations, whenever there is a human factor (i.e., any human actor involved in observing anything), variability, and hence uncertainty, can emerge~\cite{veress1993reliability},~\cite{van2016association},~\cite{braun2012agreement} as if the observers were called to observe and account for \emph{phantom} phenomena.
%FC come vuoi, se non vuoi essere didascalico togliamo pure la parentesi
%(hence the idea of ghost standard mentioned in the heading of this section). 
In all of these cases, IRR scores can give an idea of the extent the data that doctors collect, which glitter in medical datasets, are golden or alloyed.

Medical researchers use several techniques to measure IRR: the most frequently used is the Cohen's Kappa, although this is applicable only to categorical values assigned by two raters. Recently also the Krippendorf' Alpha has found some application, probably for its known advantages on the kappa, like the capability to address any number of raters (not just two), values of any level of measurement (i.e., categorical, ordinal, interval, etc.), and datasets with missing values, which are very common in medical records. 

All these metrics are intended to assess the degree of agreement \emph{beyond chance} (that is considering the fact that raters can agree not only because they believe the same thing, but also by chance). As such, there is a lot of controversy on their validity (lacking any model of how chance affects the rater decisions, let alone of the different ways to \emph{misinterpret} a phenomenon). Little consensus there is, above all, regarding how \emph{to interpret} their scores. Medical scholars usually rely on a convention from the 1970s by Landis and Koch~\cite{landis1977measurement}, where the authors indicated a \emph{fair agreement} for reliabilities between 0.21 and 0.40, \emph{moderate agreement} when its measure is between 0.41 and 0.60, \emph{substantial agreement} when between 0.61 and 0.80, and \emph{almost perfect agreement} above 0.81. This convention established over time and spread wide, although the authors themselves defined the above divisions ``clearly arbitrary'' (p. 165). More recently, Krippendorf admitted that there is no set answer to the question ``what is the acceptable level of reliability?''~\cite{krippendorff2012content} and that the answer could only ``be related to the validity requirements imposed on
the research results, specifically to the costs of drawing wrong conclusions [and whether] the analysis will affect someone's life'' (p. 325). Accordingly, and more conservatively than Landis and Koch, Krippendorf suggested to ``rely only on variables with reliabilities above .8, [\ldots] to consider variables with reliabilities between .67 and .80 \emph{only for drawing tentative conclusions} [and furthermore warned that] ``even a cutoff point of a = .80 --~meaning \emph{only} 80\% of the data are coded or transcribed to a degree better than chance~--~is a \emph{pretty low } standard by comparison to standards used in engineering, architecture, and \emph{medical research}'' (p. 325, emphasis added). As a matter of fact, medical data reach seldom such an agreement level.

To give some concrete examples of this last point, let us consider the ambit where ML-DSS have recently reached levels of diagnostic accuracy (at least) on a par with human doctors~\cite{gulshan2016development}. Even in this case, the reported IRR scores for the corresponding reference standards are quite low in light of the above recommendations. Here we mention two cases: diabetic retinopathy, which is cause of 1 case of blindness out of 10; and skin cancer, which results in approximately 80,000 deaths a year and it is the most common form of cancer in many Western countries. In the former case~\cite{gulshan2016development}, a convolutional neural network was recently trained and then evaluated in regard to the detection of diabetic retinopathy in a wide dataset of retinal fundus photographs. In this study, the authors reported high levels of both sensitivity and specificity for their ML-DSS according to the gold standard that they decided to adopt, i.e. the majority decision of a panel of board-certified ophthalmologists analyzing the same retinal fundus photographs. As a matter of fact, some authors reported that the adoption of this gold standard can be considered controversial~\cite{wong2016artificial}, since this may compare unfavorably with other gold standards used in previous studies on diabetic retinopathy (i.e. standardized centralized assessment of images or optical coherence tomography). In fact, the prevalence of diabetic retinopathy may vary significantly whether this condition is evaluated through monocular fundus photographs or, rather, through optical coherence tomography~\cite{wang2016comparison}. This could turn out to be relevant since diagnostic accuracy metrics are dependent on the prevalence of diseases according to the Bayes' theorem. Thus two questions are at stake here. On the one hand, whether similar successful results would have been obtained if a different gold standard (e.g., optical coherence tomography) had been used. On the other hand, even if we assume eye (fundus) photographs as an indisputable, unique and reliable gold standard for the diagnosis of diabetic retinopathy, IRR among opthalmic care providers has been shown to be very low (i.e., kappa between 0.27 and 0.34 for different diagnostic %RR endpoints
analyses); and still inadequate, even if higher, among retina specialists (kappa between 0.58 and 0.63)~\cite{ruamviboonsuk2006interobserver}. 
Another recent successful use of ML in diagnostics refers to the high accuracy exhibited by a convolutional neural network aimed at diagnosing skin cancers by differentiating those cases from benign skin lesions~\cite{esteva2017dermatologist}. In this case, the performance of the ML-DSS was evaluated against 21 dermatologists on biopsy-proven clinical images. Biopsy results are histological examinations, and are therefore considered the most reliable and ``standard'' gold standard. Despite this common assumption, far from optimal IRR scores have been observed among dermopathologists considering the histological diagnosis of clinically difficult cutaneous lesions, with kappa values ranging from 0.31 to 0.80 according to different diagnostic analyses~\cite{braun2012agreement}.
%RR analyses?
\section{Garbage in, Gospel out}

The question of the quality of medical record and of the data extracted from there is still understudied~\cite{stetson2012assessing,cabitza2016information}, let alone in regard to machine learning projects~\cite{feldman2017beyond}. The assumption that medical data could support secondary uses has been challenged since almost 25 years ago, and also strongly so, e.g., by Reiser~\cite{reiser1991clinical}, who described several cases of erroneous, missing and ambiguous data, and by Burnum~\cite{burnum1989misinformation}, who provocatively wrote that ``all medical record information should be regarded as suspect; much of it is fiction” (p. 484)''. Burnum even added that the introduction of health information technology had not led to improvements in the quality of medical data recorded therein, but rather to the recording of a greater quantity of ``bad data''\footnote{Moreover, Burnum traced back this lie of the land to ``standards of care and a reimbursement system [that is] blind to biologic diversity''.}. In those same years, van der Lei was among the first ones to warn against the reuse of clinical data for other objectives other than care and proposed what since then is known as the first law of informatics: `[d]ata shall be used only for the purpose for which they were collected'~\cite{van1991use}.

In light of the phenomena of both low quality and uncertainty that are intrinsic to the production of medical data\footnote{The reader should notice that low quality can be assessed only \emph{ex-post}, as information bias, by comparison with an unbiased (if any) gold standard. By assuming a certain amount of bias in medical data, and by measuring actual observer variability, available medical data can be considered uncertain \emph{ex-ante}.}, what are the main implications for the machines that are fed with this information? As widely known, many factors can contribute in downsizing the performance of a ML-DSS. Just to mention a few that we observed in the hospital domain: the fact that medical data seldom meet the common assumptions that training data should ideally possess. For instance, their attributes are seldom independent and identically distributed (IDD); their distribution is neither uniform nor normal; missing data do not occur randomly (in fact they often indicate an either good or just better health condition that relieves practitioners from the need to record it with continuous effort, see below); data can be strongly unbalanced with respect to the number of healthy and positive cases, or to the real prevalence of a pathological condition; they are not temporally stable (for instance, computer interpretations of electrocardiograms recorded just one minute apart were found significantly different in 4 of 10 cases in~\cite{spodick1997computer}); they can fall short of representing the target population (\emph{sampling bias}) or to make explicit any potential confounding variable (especially those related to ``external'' medical interventions~\cite{paxton2013developing}). 

In this view, misclassified cases by information bias could be seen as just another issue to cope with (although one of the most serious ones~\cite{frenay2014classification}). However, our point is that  considering misclassification only as a defect of data collection is a conceptual error as long as it is considered a \emph{mis}-classification: as we saw above, it should be more properly considered as a classification where independent observers disagree and classify the same phenomenon differently, to the best of their competence, perspicacity and perceptual acuity. 

This variability if often neglected even by doctors, and few studies indulge in reporting low IRR scores (also because this kind of studies is believed to disturb the doctors' morale~\cite[p. 193]{reiser1981medicine}). No wonder then that the related uncertainty is dispelled as closely as possible to the ``source'', as also the official guidelines for medical coding and reporting\footnote{(International Classification of Diseases, Ninth Revision, Clinical Modification, ICD9-CM).} ratify in an explicit way: \textit{``If the diagnosis documented at the time of discharge is qualified as `probable', `suspected', `likely', `questionable', `possible', or `still to be ruled out', or other similar terms indicating uncertainty, code the condition as if it existed or was established''}~\cite[p.90]{centers2011icd}. Alternatively uncertainty is sublimated in the (statistically significant) consensus of a sufficiently wide group of experts~\cite{svensson2015automated}.

Lastly, adopting different gold standards can affect ML-DSS significantly. We illustrate this by mentioning the case of Carpal Tunnel Syndrome (CTS): this is a kind of functional hand impairment that is frequently observed and is due to the compression of the median nerve at the wrist. This syndrome is commonly diagnosed and often referred to surgical treatment %RR cambiato un po' il seguente..
through two different gold standards whenever suspected by different specialists i.e. the sole physical examination by \emph{orthopedic surgeons} or a nerve conduction examination (electromyography) by \emph{neurologists}~\cite{bachmann2005consequences}.
In the last years, alternative diagnostic methods have been proposed to improve diagnostic accuracy for CTS, like the ultrasonography of the median nerve of the arm. These tests have shown different results in accuracy metrics when compared to the current standards mentioned above (i.e. physical examination or electromyography)~\cite{bachmann2005consequences}. These diagnostic divergences, if neglected in the training of a ML classifier aimed at the diagnosis of CTS~\cite{maravalle2015carpal}, may result in the ossification into the model of an arbitrarily partial version of the ground truth (that is whether patient X is really affected by CTS or this syndrome can be ruled out) and hence to unpredictable downstream \emph{clinical} consequences. For instance, it has been observed that a number of patients diagnosed with CTS who had undergone surgery did not receive any relevant benefit from the invasive treatment, and that this could be explained in terms of wrong upstream diagnoses~\cite{graham2006diagnosis}. 

A ML-DSS that has learned the \emph{uncertain} (i.e., right for a standard, wrong for another) mapping between the patient's features and one single diagnosis will propose its advice within a dangerous ``close-world assumption'' (that is: all the relevant features have been considered and the mapping between the input and the desired output is acquired as accurate and reliable), which is never challenged \emph{by design}. In other words, the model could suffer from \emph{algorithmic bias}, that is discriminate patients according to an arbitrary preference for a gold standard over an alternative one\footnote{As said in Section~\ref{sec:infobias}, algorithmic bias regards any assumption or heuristics that is adopted to make prediction faster or easier. In the ML literature, this latter term has recently acquired a more human-flavor, indicating the extent a classifier can reiterate or even exacerbate the typical discriminations affecting human classification or decision, like those based on gender, race, religion, health and income~\cite{hajian2016algorithmic}. Algorithmic bias has then acquired the sense of a factor explaining any ML classification that, if made by a human, would likely denote unfair inclination or prejudice. However, technically speaking what now is considered a form of algorithmic bias would not even be considered a bias according to the Mitchell's definition (see Section~\ref{sec:infobias}), as long as the basis for a discriminatory prediction is actually latent and implicit in the training data themselves. For this latter kind of algorithmic bias, the phrase ``machine bias'' could  perhaps be preferable, as in~\cite{angwin2016machine}, to avoid potentially confusing homonyms, its vagueness notwithstanding.}.

On the other hand, in the open world of hospital wards physicians are used to observer variability and less-than-perfect gold standards whenever they consult the medical data that are produced by their colleagues (and even by themselves in other work shifts, cf. the concept of intra-rater agreement). Conversely, designers of computational systems usually do not consider the case that the input of their system can be inherently and irremediably biased and inaccurate (to some extent), they assume it true. The primary concern of ML-DSS designers are the completeness, timeliness and consistency~\cite{cabitza2016information} of the datasets that they feed into the machines. There is little (if any) recognition that medical data could not be any better than ``dirty'' data with which to think to optimize a ML model adequately would be highly optimistic if not over-ambitious. Critical thinking would then suggest to look with some caution at the high accuracy rates that are often reported in the specialist literature~\cite{gulshan2016development}~\cite{esteva2017dermatologist} even assuming that model overfitting~\cite{domingos2012few} has been duly avoided. In fact, training data could be ``good'' with respect to a gold standard, but dubious according to an alternative gold standard.

This potential divergence is hardly considered when medical data are taken from the context where they have been natively produced to support coordination, knowledge sharing, sense making and decision making and they are transformed into data sets to feed in some computational systems~\cite{cabitza2016human}. 
Neglecting the gap between the primary use of medical data (i.e., care) and any secondary use (e.g., ML-DSS training) could mislead those who have to design trustworthy decision support systems, and also probably jeopardize the actual improvement of the ML-DSS performance on new and real data other than the training data. 

This points to the difference between \emph{research data}, which are usually used for ML-DSS training and optimization; and \emph{real-world} data, which are produced in real-life clinical situations. While research data are not made up on purpose to get high accuracy, they are nevertheless selected, cleansed, and \emph{engineered}~\cite{wiens2016editorial} to an extent that is completely unrealistic or unfeasible to replicate in actual clinical settings (e.g., like in~\cite{li2016integrated}). This is not only a matter of generalizability and interpretability of the model. It is also a matter of different ways to evaluate the ML-DSS performance. The most common one can be considered \emph{essentialist}~\cite{cappelletti2016appropriateness}, in that it focuses on accuracy and other performance measures (like F1-scores, and AUC-ROC) that are appraised in a laboratory setting (i.e., \emph{in laboratorio}). An alternative and still neglected approach, which we can denote as \emph{consequentialist}, focuses on the actual consequences (i.e. health outcomes) produced in situated practice (i.e., \emph{in labore}), that is in the original context of work of the physicians involved and in their actual relationship with patients, when decisions must be converted into rel-life choices that must align with the patients' attitudes, preferences, fears and hopes (as well as the economic feasibility of the available options).   

\section{Embracing uncertainty, also in computation}

As hinted above, there are many types of uncertainty in medicine, which affect medical records in different ways. For a certain attribute (i.e., variable) that is pertinent for a certain case, users could ignore what value is applicable, let alone true; or what single value is true among a finite set of values that are known to be equally applicable. Users could be uncertain between two values from the above set, or among many. Moreover, they could prefer some options with respect to others. If single users are certain about a value, they could nevertheless disagree among each other (and even with themselves over time). Ultimately, they could be uncertain among different values at various levels of confidence with respect to each other (e.g., in a dichotomous domain, which is the simplest, doctor A is \emph{fairly} certain that the condition is pathological, doctor B is {\em strongly} certain).

As shown by Svensson et al. in~\cite{svensson2015automated}, the performance of ML-DSS is negatively impacted and deeply undermined when they are fed with medical datasets that are intrinsically uncertain. Their idea is to employ conventional statistical tests to reduce the variability of the data produced by different observers by choosing the values that have been proposed by a statistically significant majority of the observers (e.g., 9 experts out of 12). However effective, this could be also seen as a way to discard the richness of a multi-value representation that accounts for a manifold phenomenon, which competent and skillful observers can describe each in her own, and partially sound yet specifically irreproachable, way. With reference to the example above, are simply 3 experts wrong, or maybe see things that the others cannot? 

Thus, if we take the ``dirtiness'' and ``manifoldness'' (seen as sides of the same coin) of medical data as a given factor of medicine, one could wonder: how can ML techniques take these constraints seriously, and even exploit them to get a richer picture of the phenomena of interest? How could ML researchers who are deeply aware of the medical complexities build models that could support human experts in their daily, and uncertain, practice?

As said in Section~\ref{sec:intro}, different kinds of uncertainty can be considered. Also, different names can be used to denote different (and similar) kinds of uncertainty. Indeed, if we look at the different classifications and taxonomies that have been proposed to describe uncertainty~\cite{Pa01}, we can find a long list of terms, like: Absence, Ambiguity, Approximation, Belief, Conflict, Confusion, Fuzziness, Imprecision, Inaccuracy, Incompleteness, Inconsistency, Incorrectness, Irrelevance, Likelihood, Non-specificity, Probability, Randomness, just to mention a few. 

Each kind of uncertainty can be addressed with specific conceptual tools to represent and manage it: probability theory, fuzzy sets, possibility theory, evidence theory, rough sets, just to mention a few also in this case. By large, the predominant role is played by probability theory, and machine learning is not an exception in this attitude. However, there exist solutions (in some case preliminary attempts) to incorporate other tools in machine learning (see for instance~\cite{Hulle11,Hulle15,Denoeux2016,Bello2017,WangZhai}). %, qualcosa di belief theory?). 
There are several reasons why these approaches are not well established in ML, as widely discussed in~\cite{Hulle15} for the fuzzy set case: sometimes the new tools are naive; there is not communication and cross-fertilization among different scholarly communities; there is a problem of credibility for many recent disciplines, which at least are not as much established as probability theory already is. 
However, if we want to deal with all of the different forms of uncertainty that we recognize affecting medical data, a specific and direct way to manage each of them, also in ML design and optimization, is preferable to a one-size-fit-all approach. 
%We believe that the above discussed flaws in data can be addressed in machine learning by making use of the appropriate tool. 
In what follows, we give some ideas on how to proceed, making reference to the biases discussed in the previous sections of the paper.
 
At first, let us consider the problem of representing a rater reliability. It is commonly assumed that ratings are \emph{exact}, though they may be classified either as ``deterministic'' or ``random''. Here ``random'' means that ``the rater is uncertain about the response category''~\cite{gwet2001}. More than a question of randomness, this description points to a form of epistemic uncertainty which can be handled by not assuming exactness, but rather introducing a sort of \emph{graduality} on the judgment scale of a rater. For instance, we could have three levels of certainty (i.e.,~low/fair/good) on the assigned rating and/or the rater can express her uncertainty by selecting more than one category with its own level of certainty. 
For instance, a physician could affirm to be sure with a good confidence that a patient suffers from disease A. Another one can be undecided between disease A and disease B with a low certainty on both.
This kind of uncertainty can be applied also to the input data (symptoms)
and we can represent the fact that a patient does {\em perhaps} suffer from headache and {\em surely} from nausea. 
This situation can be handled with Possibility Theory and, in particular, with its simplified form of certainty-based model~\cite{PP15}, which is more interpretable and simple from a computational standpoint. As an example, let us consider the data in Table~\ref{tb:possibility}. The attribute ``bicuspid aortic valve'', which is dichotomic (yes/no), is associated with a three-way expression of certainty (namely, confident/highly confident/sure). We also notice that these expressions are ordered, i.e., confident$<$highly confident$<$sure. Similarly, the attribute ``mitral rigurgitation'' is uncertain with respect to the value to be assigned to patients P5 and P6. This uncertainty is expressed through two alternative options (thus excluding all the other values).
\begin{table}[ht]
\begin{center}
\begin{tabular}{ccccc}
\hline
Patient&Mitral rigurgitation&Acute dyspnea&Bicuspid aortic valve&EKG stress test\\
\hline
P1&No&\begin{tabular}[c]{@{}c@{}} Heart failure \\(highly confident)\end{tabular}&\begin{tabular}[c]{@{}c@{}}Yes\\ (sure)\end{tabular}&\begin{tabular}[c]{@{}c@{}}Positive\\ (highly confident)\end{tabular}\\
\hline
P2&Mild&
\begin{tabular}[c]{@{}c@{}} COPD \\ (highly confident) \end{tabular} &\begin{tabular}[c]{@{}c@{}}No\\ (confident)\end{tabular}&\begin{tabular}[c]{@{}c@{}}Negative \\(highly confident)\end{tabular}\\
\hline
P3&Moderate& \begin{tabular}[c]{@{}c@{}}Pneumonia \\(highly confident)\end{tabular}&\begin{tabular}[c]{@{}c@{}}No \\(confident)\end{tabular}&\begin{tabular}[c]{@{}c@{}}Not performed\end{tabular}\\
\hline
P4&Severe&\begin{tabular}[c]{@{}c@{}}Heart failure\\(highly confident)\\or pneumonia\\(low confident)\end{tabular}&\begin{tabular}[c]{@{}c@{}}Yes \\(highly confident)\end{tabular}&\begin{tabular}[c]{@{}c@{}}Negative\\ (confident)\end{tabular}\\
\hline
P5&\begin{tabular}[c]{@{}c@{}}Moderate \\or Severe\end{tabular}&\begin{tabular}[c]{@{}c@{}}Heart failure\\(highly confident)\\or pneumonia\\(highly confident)\end{tabular}&Undetermined&Undetermined\\
% * <dvd.ciucci@gmail.com> 2017-10-05T12:56:26.141Z:
%
% ^.
\hline
P6&\begin{tabular}[c]{@{}c@{}}Mild or\\Moderate\end{tabular}&\begin{tabular}[c]{@{}c@{}}COPD or\\ pulmonary embolism \\or pneumonia\end{tabular}&Not applicable&Not available\\
\hline
P7&Undetermined&\begin{tabular}[c]{@{}c@{}}Heart failure \\and pneumonia\end{tabular}&Not applicable&\begin{tabular}[c]{@{}c@{}}Positive\\(confident)\end{tabular}\\
\hline
\end{tabular}
\caption{An exemplificatory dataset extracted from the field of work that has not undergone any post-processing and data cleansing task. Not available means a missing value that nevertheless should be represented. Not applicable denotes a structurally missing value. COPD stands for `Chronic Obstructive Lung Disease'}
\label{tb:possibility}
\end{center}
\end{table}
Of course, ML-DSS have to be modified to comply with this model, and some steps in this direction already exist~\cite{Hulle03,Haouari09,CF17}.

\begin{figure}[tb]
\centering
 \includegraphics[width=.8\textwidth]{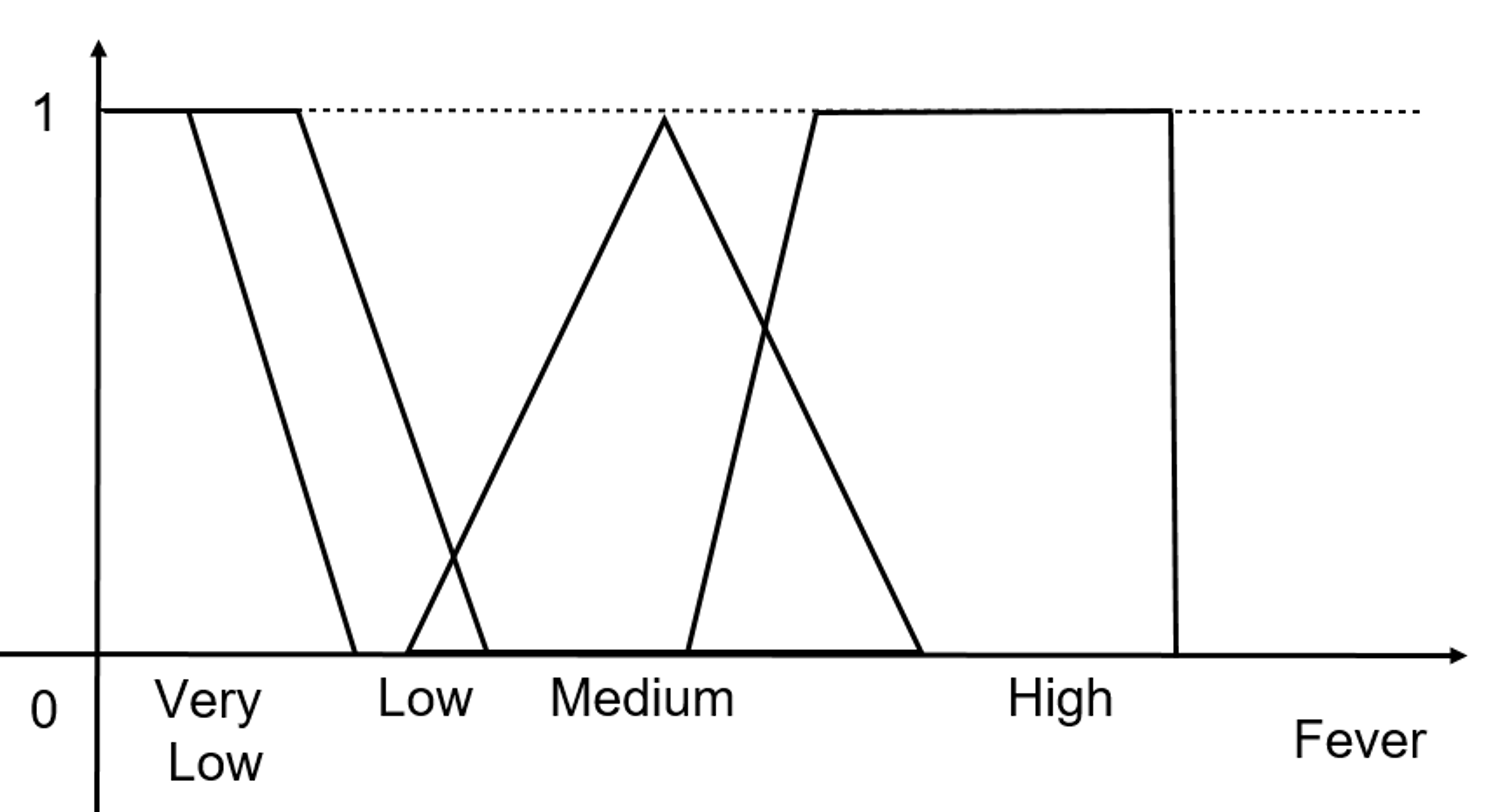}%[width=.85\textwidth]{fuzzy.png}
\caption{A typical representation of a so-called linguistic variable, fever in this case, with four values: very low, low, medium and high, each represented by a fuzzy set. The ordinate reports the measured temperature of patients, the abscissa the membership degree of patients to the four fuzzy sets. The reader should notice that the boundaries between the fuzzy sets do not necessarily have to be symmetrical and that, given an abscissa (i.e. a temperature value) the sum of the ordinates (i.e., percentages) must not necessarily equal 1.}
\label{fig:fuzzy}   
\end{figure}

A related (but different) problem is the possibility to express the nuances of a medical condition. For instance, a patient could suffer (with certainty) from {\em mild} headache and {\em strong} nausea or, with reference to Table~\ref{tb:possibility}, a {\em severe} mitral rigurgitation. We notice that in a dichotomous situation we would be forced to say {\em no} headache, {\em yes} nausea and {\em yes} mitral rigurgitation\footnote{Let us emphasize that this is different from saying that headache has a probability of $0.2$ and nausea a probability of $0.8$: probability implies the fact that we are more or less sure on a dichotomic choice, while it does not represent any actual nuance of either clinical signs or symptoms.}. Moreover, these nuances could be further differentiated by expressing the level of how much a condition is, say, low, high, and severe in terms of degrees or normalized scores. This kind of information can be modeled by fuzzy sets \cite{Hulle11,Hulle15}  as exemplified in Figure~\ref{fig:fuzzy} for the attribute fever.
 
The above two approaches of possibility theory and fuzzy sets are adequate to model uncertainty in the input data of the ML-DSS, uncertainty that can then be transferred to the output. On the other hand, {\em three-way decision} has the scope to produce an uncertain output (in front of a certain or uncertain input) \cite{Yao15,Yao16}. In this framework, objects are classified according to three levels instead of two, as in standard decision theory. In the present contest,  these three levels can have different interpretations, like ill/healthful/unknown, disease-A/disease-B/undecided. Clearly, the new introduced third level (unknown/undecided) expresses an indecision on the outcome which may be addressed by making new investigations on the patient at hand.

As a further problem, let us consider a numerical information, such as the one obtained by some measurement. Of course, any point value brings some extent of imprecision: this can be due, e.g., to the instrument itself, or related to the average value among repeated measures. Thus, we can consider to represent the information in terms of intervals and work with them natively. To this aim, interval arithmetics~\cite{MBC09} and fuzzy arithmetics~\cite{Lodwick08} give the formal instruments to operate with this kind of representation. 

%Other forms of interval, are e lower and upper probabilities of evidence theory, whose use in ML recently found 

Finally, it is well known that data often come with missing values. A standard approach in ML is to impute them in order to get a complete dataset, since several techniques require completeness to work properly. Imputation, as it is clear, ``corrupts'' the original information and some spurious values are necessarily introduced. 
On the contrary, we should not get rid of the missing values and rather consider that a missing value can have several meanings: ignorance of the real value; intrinsic non-existence; and even, as plain as it can be, that ``all is well'' (or ``nothing new happened'') and the clinicians did not deem necessary to produce a new data~\cite{de2007completeness}. An example of the ``hidden meaning'' of missing values has been recently highlighted in a study on cardiovascular risk prediction where several ML models were able to exhibit a higher predictive accuracy when compared to a traditional risk model (the American Heart Association/American College of Cardiology risk prediction algorithm) ~\cite{weng2017can}. In this study, body mass index (BMI), which was included among the variables tested for cardiovascular risk prediction, resulted in many missing values, and \emph{'missing BMI'} turned out to be a variable selected by the deep learning model for predicting a lower risk of cardiovascular events, i.e. it was selected as a protective variable. This is consistent with the post-hoc acknowledgment of the habit of many physicians \emph{not} to record the value of BMI in patients believed to be at lower global cardiovascular risk, especially when this index was likely in the normal range (i.e., the patient was not clearly overweight or obese). 
Rough set theory includes rule induction methods that work in presence of missing values and also for different meanings of it. In particular, we point the attention to the works by Gryzmala-Busse and his MLEM algorithm (see for instance~\cite{GBusse10,GBusse11}).
%------------------------
\section{Conclusions}
\epigraph{Machines can seem so accurate, so right. They can make us forget who made them, and who designed into them~--~with all the possibilities of human frailty and error~--~the programs that dictate their function. They can make us forget the hands and minds behind their creation; they can make us forget ourselves.}{\textit{Stanley Joel Reiser~\cite{reiser1984machine}}}

Fox~\cite{fox2000medical} in her relevant work on the sociology of medical knowledge once wrote that uncertainty has become the hallmark of the entire field of medicine. For this reason, confronting uncertainty has been the first and foremost driver for the introduction of computational Decision Support Systems (DSS) in medicine and their increasingly wide adoption in clinical settings~\cite{patel2009coming}. We could just speculate on why medicine has turned to technology to ``make sense of health data''\footnote{We could not exclude that this was partly due to the greater expectations of patients, who perhaps were more fatalistic in the past and now are rather consumers of health services and demanding clients of medical consultants; but, more paradoxically, reading pieces like the Nature one, it looks like technology has created the problem by making available more data to ``sift through and find answers to questions about health'' (ibidem) that now it is called to solve, e.g., with artificial intelligence softbots that crawl among million of articles to suggest the best scientific evidence available for the case at hand.} (to cite a position paper on Nature published a couple of years ago~\cite{elliott2015informatics}). Quite subtly, Katz~\cite{katz2002silent} has argued that the traditional mechanisms that physicians use to adopt to cope with uncertainty (e.g., terminological standards, standard care protocols, guidelines based on statistical studies) could have had a role in slowly pushing them towards disregarding or even opposing uncertainty.

Irrespective of the root causes of this situation, the digitization of medicine has contributed to shifting the idea of uncertainty, from being a natural and irreducible element of medical practice~\cite{rosenfeld2003uncertainty,simpkin2016tolerating} in the interpretation of subtle and sometimes contradictory clues in the existential and complex context of idiosyncratic patients, into the domain of those rational problems that can be modeled to pursue an engineering solution, or even a computational one.

In this paper we have briefly explored the blurring boundaries between what computer scientists and medical doctors pursue in medical data: data accuracy and completeness the former ones~\cite{stetson2012assessing}; trustworthiness and meaningfulness the latter ones~\cite{cabitza2016information}. We have also shed light on information bias and observer variability, which separate us from getting an absolute true, universal and reliable \emph{representation} of a physical (let alone psychological or mental) phenomenon. In particular for the ML designers, we have pointed out that information bias does not regard only the labelling of data set, i.e., the information on which a ML-DSS is trained to predict other labels accurately; but it also (and above all) affects the whole input data, in both training and prediction, especially in regard to nominal and ordinal variables. 

In light of these different viewpoints, we outline a couple of recommendations along the general framework by Domingos, who conceives ML problems as a combination of \emph{representation}, \emph{optimization} and \emph{evaluation}~\cite{domingos2012few}. From the representation perspective, computer scientists should not settle for ``polished data'' but rather ``get to the source'' of medical data: the multiple, possibly divergent opinions of experts. This means to be wary of researches where the gold standard is not reported or it is a dataset annotated by a single, or just a couple of physicians (especially if kappa or similar agreement scores are below .8). Moreover, if the adopted gold standard is based on a consensus reconciliation of divergent opinions, the authors of those researches should also be aware that they proceed considering all of these divergences as plain mistakes. If they are less than certain that this is fair, they should offer a word of caution on the potential arbitrariness of the clearly-cut classification they have used in their study. In the study design phase authors could also ask the competent observers the degree of self-perceived confidence with which they share their ideas and produce their data. This ordinal scale could be used to weight the multiple values of a single representation, so that the ML algorithms can leverage again the knowledge of the domain experts to build a coherently \emph{fuzzy} representation. Furthermore they could annotate the representativeness of each value in terms of a three-way partitioning (e.g, belonging to the majority opinion, belonging to the minority, belonging to neither with statistical significance, as done in~\cite{cabitza2017exploiting}). In any case, ML researchers should always report how they did collect their ground truth, and be explicit on what gold standard they relied on. 

In regard to optimization, further research should be devoted in transferring techniques and methods from the rough set theory (e.g.,~\cite{tsumoto2017medical}) domain into the ML arena, as seen above. In regard to evaluation, the ball could be passed to the medical practitioners again. They should develop a wariness of any essentialist evaluations of ML-DSS performance that are carried out \emph{in laboratorio} and on \emph{research data}. Rather, they should demand to the ML-DSS designers (and their advocates) \emph{evidence-based} validations of their systems~\cite{coiera2016looking}, and adopt them only once some further information has been given about, e.g., the size of the diagnostic improvement detected, the trade-off between specificity (avoiding false positive, i.e., overtesting and overtreatment) and sensitivity (avoiding false negatives, i.e., failing to treat and cure); between the internal (i.e., bias) and external (i.e., variance) validity of the model (regarding also the extent the ML-DSS could fit multimorbid cases, instead of being excessively specialized for one disease); and between its prediction power and its interpretability~\cite{caruana2015intelligible}, that is its \emph{scrutability} by doctors and lay users to understand why the ML-DSS has suggested them a certain decision over possible others~\cite{lipton2016mythos} and make the ``hybrid'' agency of man-and-machine more accountable towards the colleagues, the patients and their dears. 

Even more than that, ML-DSS should be object of a \emph{value-based} assessment, where researchers invest time and effort on the evaluation of their systems in the mid- and long-term after their deployment in real settings and their appraisal is conducted in terms of user and patient satisfaction, in terms of effect size on clinical outcomes, and eventually in terms of cost reduction or better service provision. All these elements should not be overlooked or given for granted, especially in light of the perils of automation bias (such as deskilling, technology overreliance and overdependance~\cite{goddard2012automation}) not least the surreptitious increase of trust by doctors in numbers and the ``objective facts'' (cf. McNamara fallacy) that the reckless application of \emph{machine learning} in response to an excessive \emph{human yearning} for certainty could bring in, especially in fields where this is likely to be only a dream of ignorance.

% \section*{Acknowledgement}
% This work, without the comments and insights of RR and CA, dear friends, compassionate doctors, and great diagnosticians, just could not be written. 

\bibliographystyle{splncs03}   
\bibliography{uncertainty-biblio,raffaele}  % name your BibTeX data base

\begin{thebibliography}{10}
\providecommand{\url}[1]{\texttt{#1}}
\providecommand{\urlprefix}{URL }

\bibitem{althubaiti2016information}
Althubaiti, A.: Information bias in health research: definition, pitfalls, and
  adjustment methods. Journal of multidisciplinary healthcare  9,  211 (2016)

\bibitem{andrews2007variation}
Andrews, J.E., Richesson, R.L., Krischer, J.: Variation of snomed ct coding of
  clinical research concepts among coding experts. Journal of the American
  Medical Informatics Association  14(4),  497--506 (2007)

\bibitem{angwin2016machine}
Angwin, J., Larson, J., Mattu, S., Kirchner, L.: Machine bias: There’s
  software used across the country to predict future criminals. and it’s
  biased against blacks. ProPublica, May  23 (2016)

\bibitem{bachmann2005consequences}
Bachmann, L.M., J{\"u}ni, P., Reichenbach, S., Ziswiler, H.R., Kessels, A.G.,
  V{\"o}gelin, E.: Consequences of different diagnostic ‘gold standards’ in
  test accuracy research: Carpal tunnel syndrome as an example. International
  journal of epidemiology  34(4),  953--955 (2005)

\bibitem{Bello2017}
Bello, R., Falcon, R.: Rough Sets in Machine Learning: A Review, pp. 87--118.
  Springer International Publishing, Cham (2017)

\bibitem{bowker2000sorting}
Bowker, G.C., Star, S.L.: Sorting things out: Classification and its
  consequences. MIT press (2000)

\bibitem{braun2012agreement}
Braun, R., Gutkowicz-Krusin, D., Rabinovitz, H., Cognetta, A.,
  Hofmann-Wellenhof, R., Ahlgrimm-Siess, V., Polsky, D., Oliviero, M., Kolm,
  I., Googe, P., et~al.: Agreement of dermatopathologists in the evaluation of
  clinically difficult melanocytic lesions: how golden is the ‘gold
  standard’? Dermatology  224(1),  51--58 (2012)

\bibitem{burnum1989misinformation}
Burnum, J.F.: The misinformation era: the fall of the medical record. Annals of
  Internal Medicine  110(6),  482--484 (1989)

\bibitem{cabitza2017breeding}
{Cabitza}, F.: {Breeding electric zebras in the fields of Medicine}. ArXiv
  e-prints  (2017), \url{https://arxiv.org/abs/1701.04077}

\bibitem{cabitza2016information}
Cabitza, F., Batini, C.: Information quality in healthcare. In: Data and
  Information Quality, chap.~13, pp. 421--438. Springer (2016)

\bibitem{cabitza2017exploiting}
Cabitza, F., Ciucci, D., Locoro, A.: Exploiting collective knowledge with
  three-way decision theory: Cases from the questionnaire-based research.
  International Journal of Approximate Reasoning  83,  356--370 (2017)

\bibitem{cabitza2016human}
Cabitza, F., Locoro, A.: Human-data interaction in healthcare: Acknowledging
  use-related chasms to design for a better health information. In: the Procs
  of the 8th IADIS International Conference on e-Health (2016)

\bibitem{cappelletti2016appropriateness}
Cappelletti, P.: Appropriateness of diagnostics tests. International journal of
  laboratory hematology  38(S1),  91--99 (2016)

\bibitem{carey2009blood}
Carey, I., Nightingale, C., DeWilde, S., Harris, T., Whincup, P., Cook, D.:
  Blood pressure recording bias during a period when the quality and outcomes
  framework was introduced. Journal of human hypertension  23(11),  764 (2009)

\bibitem{caruana2015intelligible}
Caruana, R., Lou, Y., Gehrke, J., Koch, P., Sturm, M., Elhadad, N.:
  Intelligible models for healthcare: Predicting pneumonia risk and hospital
  30-day readmission. In: Proceedings of the 21th ACM SIGKDD International
  Conference on Knowledge Discovery and Data Mining. pp. 1721--1730. ACM (2015)

\bibitem{chockley2016end}
Chockley, K., Emanuel, E.: The end of radiology? three threats to the future
  practice of radiology. Journal of the American College of Radiology  13(12),
  1415--1420 (2016)

\bibitem{CF17}
Ciucci, D., Forcati, I.: Certainty-based rough sets. In: Rough Sets -
  International Joint Conference, {IJCRS} 2017 Proceedings (2017), in press

\bibitem{coiera2016looking}
Coiera, E.: Looking for evidence-based medical informatics. Recenti progressi
  in medicina  107(3),  124--126 (2016)

\bibitem{darcy2016machine}
Darcy, A.M., Louie, A.K., Roberts, L.W.: Machine learning and the profession of
  medicine. Jama  315(6),  551--552 (2016)

\bibitem{Denoeux2016}
Den{\oe}ux, T., Kanjanatarakul, O.: Evidential Clustering: A Review, pp. 24--35
  (2016)

\bibitem{deo2015machine}
Deo, R.C.: Machine learning in medicine. Circulation  132(20),  1920--1930
  (2015)

\bibitem{dharmarajan2016treatment}
Dharmarajan, K., Strait, K.M., Tinetti, M.E., Lagu, T., Lindenauer, P.K., Lynn,
  J., Krukas, M.R., Ernst, F.R., Li, S.X., Krumholz, H.M.: Treatment for
  multiple acute cardiopulmonary conditions in older adults hospitalized with
  pneumonia, chronic obstructive pulmonary disease, or heart failure. Journal
  of the American Geriatrics Society  64(8),  1574--1582 (2016)

\bibitem{centers2011icd}
for Disease~Control, C., Prevention, et~al.: Icd-9-cm official guidelines for
  coding and reporting. Tech. rep., Centers for Medicare \& Medicaid Services,
  Atlanta, GA, USA (2011)

\bibitem{domingos2012few}
Domingos, P.: A few useful things to know about machine learning.
  Communications of the ACM  55(10),  78--87 (2012)

\bibitem{elliott2015informatics}
Elliott, J.H., Grimshaw, J., Altman, R., Bero, L., Goodman, S.N., Henry, D.,
  Macleod, M., Tovey, D., Tugwell, P., White, H., et~al.: Informatics: Make
  sense of health data. Nature  527,  31--32 (2015)

\bibitem{esteva2017dermatologist}
Esteva, A., Kuprel, B., Novoa, R.A., Ko, J., Swetter, S.M., Blau, H.M., Thrun,
  S.: Dermatologist-level classification of skin cancer with deep neural
  networks. Nature  542(7639),  115--118 (2017)

\bibitem{feldman2017beyond}
Feldman, K., Faust, L., Wu, X., Huang, C., Chawla, N.V.: Beyond volume: The
  impact of complex healthcare data on the machine learning pipeline. arXiv
  preprint arXiv:1706.01513  (2017)

\bibitem{fox2000medical}
Fox, R.C.: Medical uncertainty revisited. Handbook of social studies in health
  and medicine pp. 409--425 (2000)

\bibitem{frenay2014classification}
Fr{\'e}nay, B., Verleysen, M.: Classification in the presence of label noise: a
  survey. IEEE transactions on neural networks and learning systems  25(5),
  845--869 (2014)

\bibitem{gillies2017viewpoint}
Gillies, A.: Viewpoint: Embracing uncertainty. Br J Gen Pract  67(658),
  215--215 (2017)

\bibitem{goddard2012automation}
Goddard, K., Roudsari, A., Wyatt, J.C.: Automation bias: a systematic review of
  frequency, effect mediators, and mitigators. Journal of the American Medical
  Informatics Association  19(1),  121--127 (2012)

\bibitem{gordon1995evaluation}
Gordon, D.F., Desjardins, M.: Evaluation and selection of biases in machine
  learning. Machine learning  20(1),  5--22 (1995)

\bibitem{graham2006diagnosis}
Graham, B.: The diagnosis and treatment of carpal tunnel syndrome:
  Surgery—whether open or closed—works, but only if the diagnosis is right.
  BMJ: British Medical Journal  332(7556),  1463 (2006)

\bibitem{greenhalgh2013uncertainty}
Greenhalgh, T.: Uncertainty and clinical method. In: Clinical uncertainty in
  primary care, pp. 23--45. Springer (2013)

\bibitem{GBusse11}
Grzymala{-}Busse, J.W.: A comparison of some rough set approaches to mining
  symbolic data with missing attribute values. In: Kryszkiewicz, M., Rybinski,
  H., Skowron, A., Ras, Z.W. (eds.) Foundations of Intelligent Systems - 19th
  International Symposium, {ISMIS} 2011, Warsaw, Poland, June 28-30, 2011.
  Proceedings. Lecture Notes in Computer Science, vol. 6804, pp. 52--61.
  Springer (2011)

\bibitem{GBusse10}
Grzymala-Busse, J.W., Grzymala-Busse, W.J.: Handling missing attribute values.
  In: Maimon, O., Rokach, L. (eds.) Data Mining and Knowledge Discovery
  Handbook, pp. 33--51. Springer US, Boston, MA (2010)

\bibitem{gulshan2016development}
Gulshan, V., Peng, L., Coram, M., Stumpe, M.C., Wu, D., Narayanaswamy, A.,
  Venugopalan, S., Widner, K., Madams, T., Cuadros, J., et~al.: Development and
  validation of a deep learning algorithm for detection of diabetic retinopathy
  in retinal fundus photographs. JAMA  316(22),  2402--2410 (2016)

\bibitem{gwet2001}
Gwet, K.: Handbook of inter-rater reliability. STATAXIS Publishing Company
  (2001)

\bibitem{hajian2016algorithmic}
Hajian, S., Bonchi, F., Castillo, C.: Algorithmic bias: From discrimination
  discovery to fairness-aware data mining. In: Proceedings of the 22nd ACM
  SIGKDD International Conference on Knowledge Discovery and Data Mining. pp.
  2125--2126. ACM (2016)

\bibitem{Haouari09}
Haouari, B., Amor, N.B., Elouedi, Z., Mellouli, K.: Na{\"{\i}}ve possibilistic
  network classifiers. Fuzzy Sets and Systems  160(22),  3224--3238 (2009)

\bibitem{hatch2017uncertainty}
Hatch, S.: Uncertainty in medicine (2017)

\bibitem{hayes2008terminal}
Hayes, S.: Terminal digit preference occurs in pathology reporting irrespective
  of patient management implication. Journal of clinical pathology  61(9),
  1071--1072 (2008)

\bibitem{Hulle03}
H{\"{u}}llermeier, E.: Possibilistic instance-based learning. Artif. Intell.
  148(1-2),  335--383 (2003)

\bibitem{Hulle11}
H{\"{u}}llermeier, E.: Fuzzy sets in machine learning and data mining. Appl.
  Soft Comput.  11(2),  1493--1505 (2011)

\bibitem{Hulle15}
H{\"{u}}llermeier, E.: Does machine learning need fuzzy logic? Fuzzy Sets and
  Systems  281,  292--299 (2015)

\bibitem{holt2017}
Indrayan, A., Holt, M.: Concise Encyclopedia of Biostatistics for Medical
  Professionals. CRC Press (2017)

\bibitem{jewett1992potential}
Jewett, M., Bombardier, C., Caron, D., Ryan, M., Gray, R., St~Louis, E.,
  Witchell, S., Kumra, S., Psihramis, K.: Potential for inter-observer and
  intra-observer variability in x-ray review to establish stone-free rates
  after lithotripsy. The Journal of urology  147(3),  559--562 (1992)

\bibitem{jha2016adapting}
Jha, S., Topol, E.J.: Adapting to artificial intelligence: radiologists and
  pathologists as information specialists. JAMA  316(22),  2353--2354 (2016)

\bibitem{katz2002silent}
Katz, J.: The silent world of doctor and patient. JHU Press (2002)

\bibitem{kooi2017}
Kooi, T., Litjens, G.J.S., van Ginneken, B., Gubern{-}M{\'{e}}rida, A.,
  S{\'{a}}nchez, C.I., Mann, R., den Heeten, A., Karssemeijer, N.: Large scale
  deep learning for computer aided detection of mammographic lesions. Medical
  Image Analysis  35,  303--312 (2017)

\bibitem{krippendorff2012content}
Krippendorff, K.: Content analysis: An introduction to its methodology. Sage
  (2012)

\bibitem{landis1977measurement}
Landis, J.R., Koch, G.G.: The measurement of observer agreement for categorical
  data. biometrics pp. 159--174 (1977)

\bibitem{leachman2017medicine}
Leachman, S.A., Merlino, G.: Medicine: The final frontier in cancer diagnosis.
  Nature  542(7639),  36--38 (2017)

\bibitem{van1991use}
van~der Lei, J., et~al.: Use and abuse of computer-stored medical records.
  Methods Archive  30,  79--80 (1991)

\bibitem{li2016integrated}
Li, X., Liu, H., Du, X., Zhang, P., Hu, G., Xie, G., Guo, S., Xu, M., Xie, X.:
  Integrated machine learning approaches for predicting ischemic stroke and
  thromboembolism in atrial fibrillation. In: AMIA Annual Symposium
  Proceedings. vol. 2016, p. 799. American Medical Informatics Association
  (2016)

\bibitem{lipton2016mythos}
Lipton, Z.C.: The mythos of model interpretability. arXiv preprint
  arXiv:1606.03490  (2016)

\bibitem{Lodwick08}
Lodwick, W.A.: Fundamentals of Interval Analysis and Linkages to Fuzzy Set
  Theory, pp. 55--79. John Wiley \& Sons, Ltd (2008)

\bibitem{lok1998accuracy}
Lok, C.E., Morgan, C.D., Ranganathan, N.: The accuracy and interobserver
  agreement in detecting the ‘gallop sounds’ by cardiac auscultation. Chest
   114(5),  1283--1288 (1998)

\bibitem{maravalle2015carpal}
Maravalle, M., Ricca, F., Simeone, B., Spinelli, V.: Carpal tunnel syndrome
  automatic classification: electromyography vs. ultrasound imaging. TOP
  23(1),  100--123 (2015)

\bibitem{mitchell1980need}
Mitchell, T.M.: The need for biases in learning generalizations. Department of
  Computer Science, Laboratory for Computer Science Research, Rutgers Univ. New
  Jersey (1980)

\bibitem{mitchell1997machine}
Mitchell, T.M.: Machine learning. 1997. Burr Ridge, IL: McGraw Hill  45(37),
  870--877 (1997)

\bibitem{de2007completeness}
de~Mul, M., Berg, M.: Completeness of medical records in emergency trauma care
  and an it-based strategy for improvement. Medical informatics and the
  Internet in medicine  32(2),  157--167 (2007)

\bibitem{obermeyer2016predicting}
Obermeyer, Z., Emanuel, E.J.: Predicting the future—big data, machine
  learning, and clinical medicine. The New England journal of medicine
  375(13),  1216 (2016)

\bibitem{o1939gallop}
O’Farrell, P.: What is gallop rhythm? Irish Journal of Medical Science
  (1926-1967)  14(10),  729--739 (1939)

\bibitem{parasuraman2010complacency}
Parasuraman, R., Manzey, D.H.: Complacency and bias in human use of automation:
  An attentional integration. Human Factors: The Journal of the Human Factors
  and Ergonomics Society  52(3),  381--410 (2010)

\bibitem{Pa01}
Parsons, S.: Qualitative approaches for reasoning under uncertainty. The MIT
  Press, Cambridge, Massachussets (2001)

\bibitem{patel2009coming}
Patel, V.L., Shortliffe, E.H., Stefanelli, M., Szolovits, P., Berthold, M.R.,
  Bellazzi, R., Abu-Hanna, A.: The coming of age of artificial intelligence in
  medicine. Artificial intelligence in medicine  46(1),  5--17 (2009)

\bibitem{paxton2013developing}
Paxton, C., Niculescu-Mizil, A., Saria, S.: Developing predictive models using
  electronic medical records: challenges and pitfalls. In: AMIA Annual
  Symposium Proceedings. vol. 2013, p. 1109. American Medical Informatics
  Association (2013)

\bibitem{PP15}
Pivert, O., Prade, H.: A certainty-based model for uncertain databases. {IEEE}
  Trans. Fuzzy Systems  23(4),  1181--1196 (2015)

\bibitem{MBC09}
Ramon~Moore, R.B., Cloud, M.: Introduction to interval analysis. SIAM (2009)

\bibitem{reiser1991clinical}
Reiser, S.J.: The clinical record in medicine part 2: Reforming content and
  purpose. Annals of internal medicine  114(11),  980--985 (1991)

\bibitem{reiser1981medicine}
Reiser, S.J.: Medicine and the Reign of Technology. Cambridge University Press
  (1981)

\bibitem{reiser1984machine}
Reiser, S.J., Anbar, M.: The machine at the bedside: Strategies for using
  technology in patient care. Cambridge University Press (1984)

\bibitem{rosenfeld2003uncertainty}
Rosenfeld, R.M.: Uncertainty-based medicine. Otolaryngology--Head and Neck
  Surgery  128(1),  5--7 (2003)

\bibitem{ruamviboonsuk2006interobserver}
Ruamviboonsuk, P., Teerasuwanajak, K., Tiensuwan, M., Yuttitham, K., for
  Diabetic Retinopathy Study~Group, T.S., et~al.: Interobserver agreement in
  the interpretation of single-field digital fundus images for diabetic
  retinopathy screening. Ophthalmology  113(5),  826--832 (2006)

\bibitem{schwartz_artificial_1987}
Schwartz, W.B., Patil, R.S., Szolovits, P.: Artificial {Intelligence} in
  {Medicine}. New England Journal of Medicine  316(11),  685--688 (Mar 1987)

\bibitem{shafiq2016patient}
Shafiq, A., Arnold, S.V., Gosch, K., Kureshi, F., Breeding, T., Jones, P.G.,
  Beltrame, J., Spertus, J.A.: Patient and physician discordance in reporting
  symptoms of angina among stable coronary artery disease patients: Insights
  from the angina prevalence and provider evaluation of angina relief (appear)
  study. American heart journal  175,  94--100 (2016)

\bibitem{shortliffe1975model}
Shortliffe, E.H., Buchanan, B.G.: A model of inexact reasoning in medicine.
  Mathematical biosciences  23(3-4),  351--379 (1975)

\bibitem{simpkin2016tolerating}
Simpkin, A.L., Schwartzstein, R.M.: Tolerating uncertainty—the next medical
  revolution? New England Journal of Medicine  375(18),  1713--1715 (2016)

\bibitem{spodick1997computer}
Spodick, D.H., Bishop, R.L.: Computer treason: intraobserver variability of an
  electrocardiographic computer system. The American journal of cardiology
  80(1),  102--103 (1997)

\bibitem{stetson2012assessing}
Stetson, P.D., Bakken, S., Wrenn, J.O., Siegler, E.L., et~al.: Assessing
  electronic note quality using the physician documentation quality instrument
  (pdqi-9). Appl Clin Inform  3(2),  164--174 (2012)

\bibitem{svensson2015automated}
Svensson, C.M., Hubler, R., Figge, M.T.: Automated classification of
  circulating tumor cells and the impact of interobsever variability on
  classifier training and performance. Journal of immunology research  2015
  (2015)

\bibitem{throckmorton2009intraobserver}
Throckmorton, T.W., Dunn, W., Holmes, T., Kuhn, J.E.: Intraobserver and
  interobserver agreement of international classification of diseases, ninth
  revision codes in classifying shoulder instability. Journal of shoulder and
  elbow surgery  18(2),  199--203 (2009)

\bibitem{timmermans2010gold}
Timmermans, S., Berg, M.: The gold standard: The challenge of evidence-based
  medicine and standardization in health care. Temple University Press (2010)

\bibitem{tsumoto2017medical}
Tsumoto, S.: Medical diagnosis: Rough set view. In: Thriving Rough Sets, pp.
  139--156. Springer (2017)

\bibitem{van2016association}
Van~Driest, S.L., Wells, Q.S., Stallings, S., Bush, W.S., Gordon, A.,
  Nickerson, D.A., Kim, J.H., Crosslin, D.R., Jarvik, G.P., Carrell, D.S.,
  et~al.: Association of arrhythmia-related genetic variants with phenotypes
  documented in electronic medical records. Jama  315(1),  47--57 (2016)

\bibitem{veress1993reliability}
Veress, B., Gadaleanu, V., Nennesmo, I., Wikstr{\"o}m, B.: The reliability of
  autopsy diagnostics: inter-observer variation between pathologists, a
  preliminary report. International Journal for Quality in Health Care  5(4),
  333--337 (1993)

\bibitem{WangZhai}
Wang, X., Zhai, J.: Learning with uncertainty. CRC Press (2017)

\bibitem{wang2016comparison}
Wang, Y.T., Tadarati, M., Wolfson, Y., Bressler, S.B., Bressler, N.M.:
  Comparison of prevalence of diabetic macular edema based on monocular fundus
  photography vs optical coherence tomography. JAMA ophthalmology  134(2),
  222--228 (2016)

\bibitem{weng2017can}
Weng, S.F., Reps, J., Kai, J., Garibaldi, J.M., Qureshi, N.: Can
  machine-learning improve cardiovascular risk prediction using routine
  clinical data? PloS one  12(4),  e0174944 (2017)

\bibitem{wiens2016editorial}
Wiens, J., Wallace, B.C.: Editorial: special issue on machine learning for
  health and medicine. Machine Learning  102(3),  305--307 (2016)

\bibitem{wong2016artificial}
Wong, T.Y., Bressler, N.M.: Artificial intelligence with deep learning
  technology looks into diabetic retinopathy screening. JAMA  316(22),
  2366--2367 (2016)

\bibitem{wu2016classifier}
Wu, H., Deng, Z., Zhang, B., Liu, Q., Chen, J.: Classifier model based on
  machine learning algorithms: Application to differential diagnosis of
  suspicious thyroid nodules via sonography. American Journal of Roentgenology
  207(4),  859--864 (2016)

\bibitem{Yao15}
Yao, Y.: Rough sets and three-way decisions. In: Ciucci, D., Wang, G., Mitra,
  S., Wu, W. (eds.) Rough Sets and Knowledge Technology - 10th International
  Conference, {RSKT} 2015, held as part of the International Joint Conference
  on Rough Sets, {IJCRS} 2015, Tianjin, China, November 20-23, 2015,
  Proceedings. Lecture Notes in Computer Science, vol. 9436, pp. 62--73.
  Springer (2015)

\bibitem{Yao16}
Yao, Y.: Three-way decisions and cognitive computing. Cognitive Computation
  8(4),  543--554 (2016)

\end{thebibliography}
% \begin{thebibliography}{5}
% %
% \bibitem {clar:eke}
% Clarke, F., Ekeland, I.:
% Nonlinear oscillations and
% boundary-value problems for Hamiltonian systems.
% Arch. Rat. Mech. Anal. 78, 315--333 (1982)

% \bibitem {clar:eke:2}
% Clarke, F., Ekeland, I.:
% Solutions p\'{e}riodiques, du
% p\'{e}riode donn\'{e}e, des \'{e}quations hamiltoniennes.
% Note CRAS Paris 287, 1013--1015 (1978)

% \bibitem {mich:tar}
% Michalek, R., Tarantello, G.:
% Subharmonic solutions with prescribed minimal
% period for nonautonomous Hamiltonian systems.
% J. Diff. Eq. 72, 28--55 (1988)

% \bibitem {tar}
% Tarantello, G.:
% Subharmonic solutions for Hamiltonian
% systems via a $\bbbz_{p}$ pseudoindex theory.
% Annali di Matematica Pura (to appear)

% \bibitem {rab}
% Rabinowitz, P.:
% On subharmonic solutions of a Hamiltonian system.
% Comm. Pure Appl. Math. 33, 609--633 (1980)

% \end{thebibliography}

\end{document}